\documentclass{article}

\usepackage{arxiv}

\usepackage[utf8]{inputenc} 
\usepackage[T1]{fontenc}    
\usepackage{hyperref}       
\usepackage{url}            
\usepackage{booktabs}       
\usepackage{amsfonts}       
\usepackage{nicefrac}       
\usepackage{microtype}      
\usepackage{cleveref}       
\usepackage{lipsum}         
\usepackage{graphicx}
\usepackage{doi}

\usepackage{subcaption}
\usepackage{microtype}
\usepackage{multirow}
\usepackage{makecell}
\usepackage{caption}
\usepackage{hyperref}

\title{HS-BAN: A Benchmark Dataset of Social Media Comments for Hate Speech Detection in Bangla}

\author{ 
    Nauros Romim, Mosahed Ahmed\\
	Dept. of Electrical \& Electronic Engineering\\
	Shahjalal University of Science and Technology\\
	Sylhet-3114 \\
	\texttt{naurosromim@gmail.com, mosahed32@student.sust.edu} \\
	\AND
	Md. Saiful Islam, \href{https://orcid.org/0000-0002-0407-6526}{\includegraphics[scale=0.06]{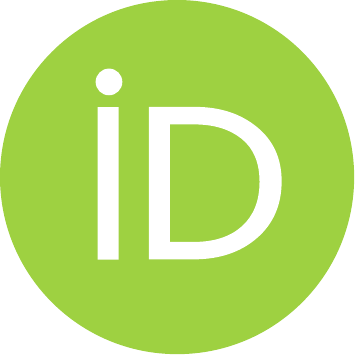}\hspace{.5mm}Arnab Sen Sharma} \\
	Dept. of Computer Science and Engineering\\
	Shahjalal University of Science and Technology\\
	Sylhet-3114 \\
	\texttt{saiful-cse@sust.edu, arnab-cse@sust.edu} \\
	\And
	Hriteshwar Talukder\\
	Dept. of Electrical \& Electronic Engineering\\
	Shahjalal University of Science and Technology\\
	Sylhet-3114 \\
	\texttt{hriteshwar-eee@sust.edu} \\
	\And
	\href{https://orcid.org/0000-0001-6540-3415}{\includegraphics[scale=0.06]{orcid.pdf}\hspace{.5mm}Mohammad Ruhul Amin} \\
	Computer and Information Science\\
	Fordham University\\
	New York, USA \\
	\texttt{mamin17@fordham.edu}
}

\renewcommand{\shorttitle}{HS-BAN}

\hypersetup{
    pdftitle={HS-BAN},
    pdfsubject={q-bio.NC, q-bio.QM},
    pdfauthor={Arnab Sen Sharma},
    pdfkeywords={First keyword, Second keyword, More},
}

\begin{document}
\maketitle

\begin{abstract}
In this paper, we present HS-BAN, a binary class hate speech (HS) dataset in Bangla language consisting of more than 50,000 labeled comments, including 40.17\% hate and rest are non hate speech.
While preparing the dataset a strict and detailed annotation guideline was followed to reduce human annotation bias. 
The HS dataset was also preprocessed linguistically to extract different types of slang currently people write using symbols, acronyms, or alternative spellings. These slang words were further categorized into traditional and non-traditional slang lists and included in the results of this paper.
We explored traditional linguistic features and neural network-based methods to develop a benchmark system for hate speech detection for the Bangla language. 
Our experimental results show that existing word embedding models trained with informal texts perform better than those trained with formal text. Our benchmark shows that a Bi-LSTM model on top of the FastText informal word embedding achieved 86.78\% F1-score. We will make the dataset available for public use.
\end{abstract}


\section{Introduction}
Hate speech (HS) has been rapidly spreading in recent years through various social media sites. HS detection in the user's comment section remains a difficult challenge due to lack of formal language syntax, spelling mistakes, and use of various slang and non-standard acronyms.\cite{nobata2016abusive}. Researchers have started working in this very challenging domain, but much of their efforts concentrated on English language\cite{schmidt2017survey}. 
There are almost 46 million Facebook\footnote{https://www.statista.com/statistics/268136/top-15-countries-based-on-number-of-facebook-users/} and 29 million Youtube\footnote{https://www.statista.com/forecasts/1146236/youtube-users-in-bangladesh} Bangladeshi users but 
there has been a severe lack of large, linguistically diverse Bangla HS datasets.

Existing datasets for Bangla HS have some significant drawbacks. First, as \textbf{Table \ref{tab: bangla HS dataset papers}} of \textit{Appendix A} shows, the datasets are not large enough. Most of them contain less than 10k sentences. Secondly, the majority portions of those datasets come from one or two domains, which makes them domain-dependent \cite{Ishmam2019_FB_pages}, \cite{Emon2019} and \cite{Chakraborty_2019}. Thirdly, annotating HS is inherently a complex and challenging task\cite{nobata2016abusive}. \cite{waseem2016you} showed how annotator's bias on HS can influence the annotation process and subsequently affect classification task. To combat this, \cite{de2018hate} created a stringent and detailed annotation guideline, providing specific points on what constitutes HS and what not. Although \cite{romim2021hate} followed a guideline to prepare a small Bangla HS dataset, it lacks inter-annotator scores, which makes it difficult to judge annotation quality. To the best of our knowledge, only \cite{karim2020deephateexplainer} mentioned Cohen's kappa score in their paper.

\begin{figure}[h!]
    \centering
    \includegraphics[height=4.5cm]{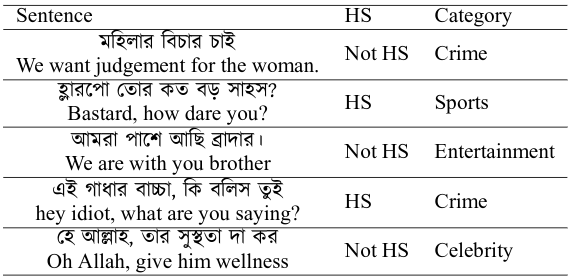}
    \caption* {Table 1: Examples of the annotate HS-BAN dataset}
    \label{tab:sample dataset}
\end{figure}

In this paper, we manifest HS-BAN, the largest binary class Bangla HS dataset comprised of more than 50,000 comments crawled from Facebook and YouTube. We present a few examples from the dataset in \textbf{Table \hyperref[tab:sample dataset]{1}}. 
To ensure cross-domain generalization, we collected comments from seven categories as presented in \textbf{Section 2}. A strict annotation guideline was followed to reduce human annotation bias for labeling the dataset achieving an inter-annotator agreement score of 0.658. 
In \textbf{Section 3}, we present the different natural language processing approaches we used to extract linguistic features by processing noisy social media text. 
Finally, in \textbf{Section 4}, we present the benchmarking outcome for hate speech classification showing that a Bi-LSTM model on top of the FastText informal word embedding achieved 86.78\% F1-score. The dataset and source code will be made publicly available to foster future research.

\section{Dataset}

\setcounter{table}{1}

\paragraph{Data Collection:}
We collected more than 100,000 comments from different YouTube channels and Facebook pages in seven categories: \textit{sports, entertainment, crime, politics, religion, celebrity}, and \textit{miscellaneous}. For each category except \textit{miscellaneous}, we compiled a list of controversial events that happened recently (2017-2020) in Bangladesh that falls under those categories. 
Then public comments were extracted on those issues by using an open-source tool called Facepager\footnote{\url{https://github.com/strohne/Facepager}}. Furthermore, for the \textit{miscellaneous} category, we searched for videos on YouTube related to Bangla TikTok, roasting, or similar videos as the comment section in these videos tend to be very toxic. 
We removed duplicate, highly similar comments based on \textit{Jaccard Index > 0.8} to reduce repetitiveness and ensure a diverse vocabulary. 

\paragraph{Annotation:} We prepared a detailed annotation guideline based on the community guidelines of Facebook and YouTube (\textit{Appendix B}). 
We worked with 50 annotators who are undergraduate students within the age range from 20 to 25 years.
They know about the sensitive nature of the task and they willingly volunteered in this research.
Three annotators tagged each comment, and the majority vote decided the final decision. The inter-annotator agreement score \cite{fleiss1971measuring} of 0.658 indicates that our dataset has moderate agreement among annotators.

\paragraph{Dataset Statistics:} 
Our dataset has $50,314$ comments in total; among them, 20,209 of the dataset comments are HS and 30,105 are Not-HS, which implies that our dataset is moderately imbalanced. A more detailed statistics of our dataset is presented in \textbf{Table \ref{tab:category word count}}. We can see that each category is also moderately imbalanced, especially the \textit{politics} and \textit{celebrity} categories, whereas crime has the highest hate speech percentage. Observing the average word count (AWC) for each category, we can find that, \textit{celebrity} category has the highest AWC per sentence with the most standard deviation while \textit{miscellaneous} has the lowest.

\begin{table}[t]
    \centering
    \resizebox{.7\textwidth}{!}{%
    \begin{tabular}{lllllll}
    
    \hline
      Category & Total  &  HS\% & AWC & Vocabulary\\
      \hline
        Sports & 5937 & 40.02 & $13.51\pm 18.27$  &16299 \\
        Entertainment & 6843 & 41.31 & $12.30 \pm 13.31$ &16915 \\
        Crime & 4969 & 43.71 &	$16.48 \pm 21.53$ &15272	\\
        Religion & 4978 & 38.31 & $13.25 \pm 21.54$	 &13717	\\
        Politics & 2665 & 33.06 & $12.90 \pm 17.93$	 &9059	\\
        Celebrity & 2394 & 30.49 & $22.96 \pm 33.59$	 &14220 \\
        Miscellaneous & 22,528 & 41.35 & $11.35 \pm 14.16$	 &33646	\\
        \hline
        Total &50314 &40.17 & $13.07 \pm 17.97$ &72660 \\
        \hline
    \end{tabular}%
    }
    
    \vspace{.3cm}
    \caption{ Category-wise Comment distribution}
    \label{tab:category word count}
\end{table}

\section{Experimental Setup}
We removed all hashtags, numbers, and non-Bangla words. However, we kept emoji and punctuation (\textit{e\&p}) in order to confirm \cite{nobata2016abusive}'s claim whether \textit{e\&p} can be good indicators for HS. 
F1 score and Matthews correlation coefficient (MCC) were used to evaluate all models. We split the dataset into train(80\%) and test(20\%) datasets using stratified sampling so that each category in both dataset contains an equal ratio of \textit{hate} to \textit{not-hate} comments. The training dataset was further divided into train(80\%) and development(20\%) datasets for feature selection and hyper-parameter tuning purposes.

\paragraph{Machine Learning Models:} We trained linear Support Vector Machine (SVM)\cite{cortes1995support} with Term Frequency Inverse Document Frequency (TF-IDF) weighted score as a feature. Regularizer C, penalty, and loss were fine-tuned to find the best hyperparameter combination. And for the deep learning model, we fine-tuned Convolutional Neural Networks (CNN) as described in \cite{zhang2015sensitivity} to get the best F1 score in the development set. We also implemented Bi-directional Long Short Term Memory (Bi-LSTM) as described in \cite{rasooli2018cross}. 

\paragraph{Word Embedding Models:} We experimented with \textit{BengFastText(BFT)} \cite{Karim2020} pretrained on 250 million Bangla articles and \textit{FastText}\cite{grave2018learning}, a multilingual model pretrained on 157 languages. Note that both are pretrained on formal texts; thus denoted as \textit{formal embedding}. Additionally, 1.47 million Bangla comments from Facebook and YouTube on eight different categories: education, entertainment, health, influencer, religion, politics, sports, technology were collected. Then four different word-embedding models were trained using those datasets. Two word2vec\cite{mikolov2013distributed} models denoted as \textit{W2V(SG)} (Word2Vec skip-gram) and \textit{W2V(CBOW)} (Word2Vec continuous bag of words) and two FastText embeddings denoted as \textit{FT(SG)} (FastText skip-gram) and \textit{FT(CBOW)} (FastText continuous bag of words) were trained using informal texts, hence denoted as \textit{informal embedding.}

\paragraph{Transformer Models:} We experimented with some transformer-based models: monolingual Bangla-BERT\cite{Sagor_2020}, multilingual BERT-cased and uncased\cite{DBLP:journals/corr/abs-1810-04805}.   
The hyperparameters were set according to literature \cite{karim2020deephateexplainer} except for mBERT-uncased. In this case, the learning rate was changed from its original value of 5e-5 to 3e-5 as the validation loss during the initial training phase was terrible. 

\section{Result and Analysis} 
\paragraph{Baseline analysis:}We present our experimental findings in table \ref{tab:result comparison}. We experimented with different combinations of char and word ngram, and only the most notable combinations were presented. It becomes apparent that in our dataset, \textit{e\&p} do not serve as valuable features. Word n-gram with \textit{e\&p} had the worst F1 score, and for char n-gram (1,6) and (2,6), when \textit{e\&p} are present, the F1 score drops. For this reason, in hyper-parameter tuning and neural network experiments, \textit{e\&p} were removed in the prepossessing step. The experimental results show that char n-gram as a feature performs better than word n-gram. Our dataset contains many spelling mistakes, which means the same words can have inconsistent and multiple spelling variations for which SVM peforms poorly \cite{schmidt2017survey}. The performance drops with experiments combining char n-gram with word n-gram further validated the fact. Overall, char ngram (1,6) is the best performing feature with F1 score of 85.86.

\paragraph{Formal vs Informal Embeddings:} We observed that, SVM with char (1-6) gram as feature outperformed neural networks with \textit{FastText} and \textit{BFT} as embedding by a good margin. \cite{kar-etal-2020-multiview} found similar observations for their informal sentiment analysis dataset. They proposed that pre-trained models based on formal text do not perform well on the informal dataset. Two FastText informal embedding \textit{FT(SG)} and \textit{FT(CBOW)} outperformed SVM in three out of four instances. Only \textit{CNN+FT(CBOW)} achieved a lower F1 score. Most notably, they were trained on a small 1.47 million informal sentences compared to embeddings such as \textit{BFT} and \textit{FastText} but still managed to outperform the bigger embeddings by a large margin. From the experiments, we conclude, word embedding trained from scratch on informal corpus will perform better for the informal dataset collected from social media.

\textit{FT(SG)} and \textit{FT(CBOW)} performed comparatively better than both \textit{W2V(SG)} and \textit{W2V(CBOW)} as presented in \textbf{Table \ref{tab:result comparison}}. According to \cite{joulin2016fasttext}, FastText is generally a better choice compared to Word2Vec for our dataset. 
{We can also see that \textit{W2V(SG)} and \textit{FT(SG)} outperformed \textit{W2V(CBOW)} and \textit{FT(CBOW)}, respectively. From this, we can summarize that word embedding created on the \textit{SG} method tends to perform better than the \textit{CBOW} method for our dataset.} Overall, combining \textit{Bi-LSTM} with \textit{FT(SG)} or \textit{Bi-LSTM+FT(SG)} achieved the best result with an F1 score of 86.85\%.

\paragraph{Transformer vs Other:} All BERT variants performed significantly worse compared to other models. Possible reasons could be that mBERT is under-tuned for Bangla informal text as it was trained with Bangla Wikipedia for which we observe the lower performance. The experimental success with informal word embedding over formal embedding support the last assumption.


\begin{table}[h!]
    
    \centering
    
    \begin{tabular}{l|l|l|l}
    \hline
         & Model & F1 & MCC\\
    \hline
    \multirow{12}{*}{Baseline}&Unigram (U) & 82.89 & 72.95 \\
             & U+e\&p &82.38 &72.17 \\
             & Bigram (B) & 63.12 & 49.23   \\
             &Trigram (T) & 24.85 & 23.60   \\
             & U+B        & 82.11 & 71.32 \\
             & U+B+T      & 81.16 & 70.38 \\
             & Char 1-6 gram(C(1,6)) &\textbf{85.97} & 77.15  \\
             & C(1,6)+e\&p &85.86 & 77  \\
    
             & C(2,6) &85.89 & 77.04   \\
             & C(2,6)+e\&p &85.76 & 76.86   \\
    
             & U+B+C(1,6) &84.71  & 75.04  \\
             & U+B+C(2,6) &84.73 & 75.10   \\
\hline
   
    \multirow{4}{*}{Formal embedding} & CNN+BFT & 71.58 & 53.06\\
                  & Bi-LSTM+BFT & 76.09 & 61.51\\
    
                  & CNN+FastText & 84 & 72.89\\
                  & Bi-LSTM+FastText &83.89 &73.43\\

    \hline
    \multirow{8}{*}{Informal embedding} & CNN+W2V(CBOW) & 80.03 & 68.42\\
                  & Bi-LSTM+W2V(CBOW) & 80.48 & 69.46\\
                  & CNN+W2V(SG) & 81.07 & 69.23\\
                  & Bi-LSTM+W2V(SG) & 80.93 & 69.25\\
                  & CNN+FT(CBOW) & 85.05 & 75.15\\
                  & Bi-LSTM+FT(CBOW) & 86.73 & 77.76\\
                  & CNN+FT(SG) & 86.58 & 77.33\\
                  & \textbf{Bi-LSTM+FT(SG)} & \textbf{86.85} & \textbf{77.90}\\
    \hline
    \multirow{4}{*}{Transformer} & mBERT-cased &81.66 &72.81\\
                 & mBERT-uncased &82.56 &73.33\\
                 & Bangla-BERT &83.68 &74.63\\
    \hline
    \end{tabular}
    
    \vspace{.3cm}
    \caption{Benchmarking the hate speech classification models.}
    
    \label{tab:result comparison}
\end{table}

\begin{table}
    \centering
    \resizebox{.7\columnwidth}{!}{%
    \begin{tabular}{l|l|l|l|l}
    \hline
      Category & Total & HS\% & F1 avg & F1 improved\\
        \hline
        Sports &4748 &40.01 &84.99 &87.70\\
        Entertainment &5468 &41.32 &87.09 &88.51\\
        Crime &3973 &43.71 &92.27 &92.77\\
        Religion &3978 &38.31 &85.92 &88.67\\
        Politics &2137 &33.04 &80.69 &86.39\\
        Celebrity &1909 &30.490 &78.84 &83.02\\
        Miscellaneous &18,021 &41.35 &85.99 &88.00\\
        \hline
    \end{tabular}
    }
    \vspace{.3cm}
    \caption{Category wise result comparison}
    \label{tab:category wise}
\end{table}

\paragraph{Category-wise Result:} 
From Table \ref{tab:category wise} we can see that for each category, the F1 scores of 5 top-performing models: SVM, Bi-LSTM+FT(SG), CNN+FT(SG), Bi-LSTM+FT(CBOW), and CNN+FT(CBOW) were averaged and presented in the \textit{F1 avg} column. \textit{Politics} and \textit{Celebrity} categories achieved the lowest average F1 scores. Two reasons might contribute to this: both categories have the lowest training data and the lowest HS\%. To check this hypothesis, we dropped some NHS in each category so that each of them has an equal number of HS and NHS. Then, the average F1 score for all five models was calculated and shown in the \textit{F1 improved} column. Note that F1 scores for \textit{Politics} and \textit{Celebrity} improved the most.
On the other hand, \textit{Crime} has the best HS\% and achieved the most \textit{F1 avg score}. And so, it also improved the least (only 0.5) after balancing.

\paragraph{Influence of slang words:} The dataset vocabulary was extracted, and slang words were annotated with the help of a linguistic expert. These slang words were divided into two types: traditional slang (TS) and non-traditional slang (NTS). NTS is not usually used as slang words but can be interpreted as such depending on the context. \textbf{Figure \ref{fig: trad-no-trad}} in \textit{Appendix C} shows some examples of both types of slang words. Then we looked at how models predicted comments that contain at least one such slang word. 
For simplicity, we ignored comments that contained both TS and NTS words.  Our initial hypothesis was that models would face more trouble detecting sentences NTS as these words are often used as HS as well as NHS. \textbf{Table \ref{tab:slang word}} shows this result. \textit{TS acc} shows how models predicted comments containing TS and \textit{NTS acc} is for NTS. We can see that models performed better when comments contained TS words rather than NTS. NTS words are often vague in meaning, and so they pose special difficulty for the models.

\label{sec:slang words}

\begin{table}[h]
    \centering
    \begin{tabular}{l|c|l}
    \hline
    Model &TS acc &NTS acc\\
    \hline
      SVM  &  84.27 &79.53\\
      Bi-LSTM+FT(SG)   &84.87 &79.40\\
      Bi-LSTM+FT(CBOW) &85.08 &77.97\\
      CNN+FT(SG) &84.95 &77.59\\
      CNN+FT(CBOW) &83.39 &74.87\\
     \hline
    \end{tabular}
    \vspace{0.3cm}
    \caption{Influence of traditional and non traditional slang words}
    \label{tab:slang word}
\end{table}

\section{Conclusion}
In this paper, we present HS-BN, the largest Bangla HS dataset collected from social media comments. 
We followed various schemes to ensure linguistic diversity and reduce repetitiveness. 
We found that emoji and punctuation do not affect HS detection. We also found that word embedding trained on informal text outperforms available embedding trained on formal text. 
We also observed that traditional slangs make the detection of hate speech much easier than that of comments with non-traditional slangs.
In the future, we plan to research on the interpretability and generalizability of hate speech detection model.

\bibliographystyle{unsrt}
\bibliography{references}  






\newpage

\section*{Appendix}
\setcounter{table}{0}
\renewcommand{\thetable}{A.\arabic{table}}

\section*{\textbf{A:} Bangla HS dataset comparision}
\begin{table}[h!]
    \centering
    \begin{tabular}{lllll}
    \hline
     & Dataset size & No of HS & \makecell{Annotation\\ guideline} &\makecell{Agreement\\ score}\\
     \hline
    (Chakraborty and Seddiqui, 2019) \cite{Chakraborty_2019} & 5,644  & 2500 & No &No\\
    (Emon et al., 2019) \cite{Emon2019} & 4,700 & 3137 & No &No\\
    (Awal et al., 2018) \cite{Awal2018} & 2665 & 1214 & No &No\\
    (Ishmam and Sharmin, 2019) \cite{Ishmam2019_FB_pages} & 5,126 & 3178 & No &No\\
    (Banik, 2019)  \cite{Banik2019} & 10,219 & 4255 & No &No\\
    (Karim et al., 2020a) \cite{karim2020deephateexplainer} & 6115 & 6115 & No &Yes\\
    (Romim et al., 2021) \cite{romim2021hate} & 30,000 &10,000 &Yes &No\\
    \textbf{Our} & \textbf{50,314} & \textbf{20,209} &\textbf{Yes} &Yes\\
    \hline
    \end{tabular}
    \vspace{0.3cm}
   \caption{Overview of binary hate speech dataset of Bangla social media}
    \label{tab: bangla HS dataset papers}
\end{table}

\section*{\textbf{B:} Annotation Guidelines}
We chose three criteria for a comment to be labeled as HS. These are: 

\begin{itemize}
    \item Deliberate attack
    \item Directed towards a specific group of people
    \item Motivated by an aspect of groups identity
\end{itemize} 

But bellow we also present a more nuanced guideline for annotation, based on the community standard of Facebook and YouTube.
\\
\\
\textbf{Criteria for HS}
    
        
    
    
        
    
        
    
        
        
        

\begin{figure}[h!]
    \centering
    \includegraphics[width=.95\textwidth]{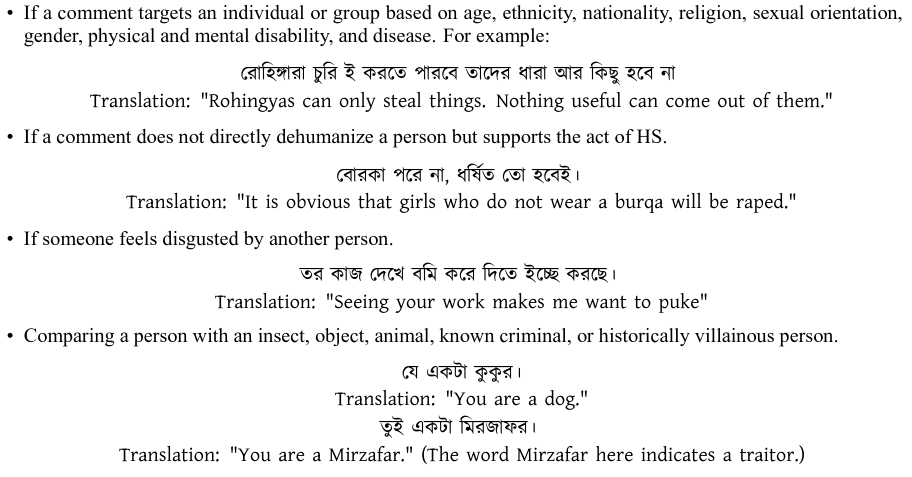}
\end{figure}

\newpage
\textbf{Criteria for not HS}

        

        
    

\begin{figure}[h!]
    \centering
    \includegraphics[width=.95\textwidth]{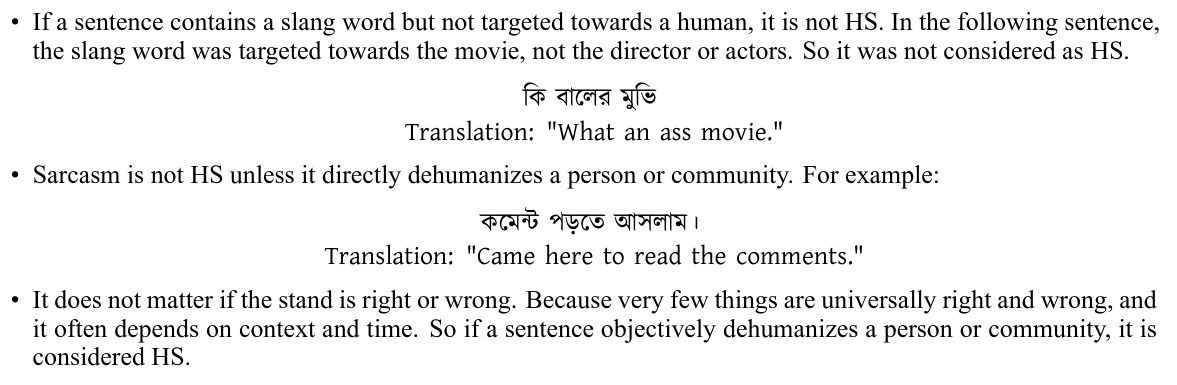}
\end{figure}

\renewcommand{\thefigure}{C.\arabic{figure}}

\section*{\textbf{C:} Traditional and non traditional slang words}
\label{sec:slang words}

\setcounter{figure}{0}

\begin{figure}[h!]
    \centering
    \begin{subfigure}[t]{0.5\textwidth}
        \centering
        \includegraphics[height=4cm]{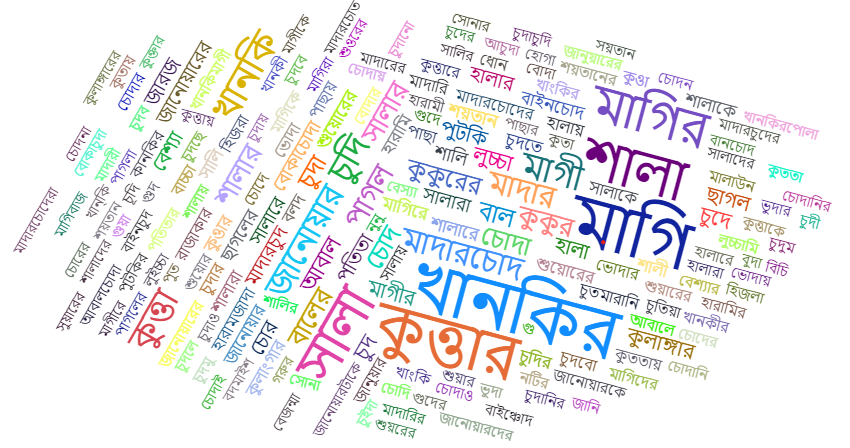}
        \caption{Traditional Slang WordCloud}
        \label{fig:trad slang}
    \end{subfigure}%
    \begin{subfigure}[t]{0.5\textwidth}
        \centering
        \includegraphics[height=4cm]{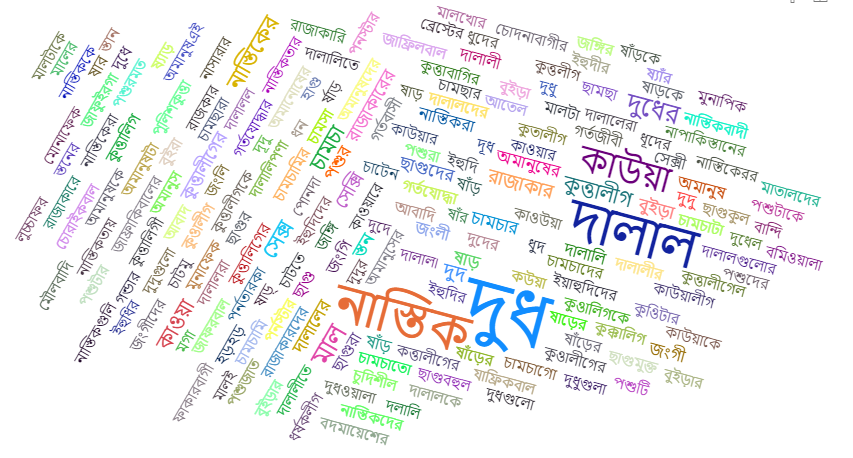}
        \caption{Non Traditional Slang WordCloud}
        \label{fig:nontrad slang}
    \end{subfigure}
    \caption{Comparison of traditional and non-traditional slang words.}
    \label{fig: trad-no-trad}
\end{figure}

\setcounter{table}{0}
\renewcommand{\thetable}{D.\arabic{table}}

\end{document}